# ROS-ELM: A Robust Online Sequential Extreme Learning Machine for Big Data Analytics[*]

Yang Liu[a], Bo He[a,*], Diya Dong[a], Yue Shen[a], Tianhong Yan[b], Rui Nian[a], Amaury Lendasse[c]

[a] School of Information Science and Engineering, Ocean University of China, 238 Songling Road, Qingdao 266100, China

[b] School of Mechanical and Electrical Engineering, China Jiliang University, 258 Xueyuan Street, Xiasha High-Edu Park, Hangzhou 310018, China

[c] Department of Mechanical and Industrial Engineering and the Iowa Informatics Initiative, 3131 Seamans Center, The University of Iowa, Iowa City, IA 52242-1527, USA

**Abstract.** In this paper, a robust online sequential extreme learning machine (ROS-ELM) is proposed. It is based on the original OS-ELM with an adaptive selective ensemble framework. Two novel insights are proposed in this paper. First, a novel selective ensemble algorithm referred to as particle swarm optimization selective ensemble (PSOSEN) is proposed. Noting that PSOSEN is a general selective ensemble method which is applicable to any learning algorithms, including batch learning and online learning. Second, an adaptive selective ensemble framework for online learning is designed to balance the robustness and complexity of the algorithm. Experiments for both regression and classification problems with UCI data sets are carried out. Comparisons between OS-ELM, simple ensemble OS-ELM (EOS-ELM) and the proposed ROS-ELM empirically show that ROS-ELM significantly improves the robustness and stability.

**Keywords:** extreme learning machine, online learning, selective ensemble, PSOSEN, robustness

---

[*] Corresponding author

Email address: `bhe@ouc.edu.cn`(Bo He).

# 1    Introduction

Due to the advancement of data acquisition, the amount of information in many fields of sciences increases very rapidly. The world is entering the age of big data. Large data set is helpful to analyze various phenomena because abundant information is available. However, it also raises many new problems. First, the computational time for big data analytics increasing rapidly. Second, various data sets require more robust learning algorithms.

Feedforward neural networks is one of the most prevailing neural networks, which is very popular for data processing in the past decades [1-2]. However, all the parameters in the networks need to be tuned iteratively. Moreover, the slow gradient descent based learning methods are always used to train the networks [3]. Therefore, the learning speed of the feedforward neural networks is very slow, which limits its applications.

Recently, an original algorithm designed for single hidden layer feedforward neural networks (SLFNs) named extreme learning machine (ELM) was proposed by Huang et al.[4]. ELM is a tuning free algorithm for it randomly selects the input weights and biases of the hidden nodes instead of learning these parameters. And also, the output weights of the network are then analytically determined. ELM proves to be a few orders faster than traditional learning algorithms and obtains better generalization performance as well. It lets the fast and accurate big data analytics becomes possible and has been applied to many fields [5-7].

However, the algorithms mentioned above need all the training data available to build the model, which is referred to as batch learning. In many industrial applications, it is very common that the training data can only be obtained one by one or chunk by chunk. If batch learning algorithms are performed each time new training data is available, the learning process will be very time consuming. Hence online sequential learning is necessary for many real world applications.

An online sequential extreme learning machine is then proposed by Liang et al.[8]. OS-ELM can learn the sequential training observations online at arbitrary length (one by one or chunk by chunk). New arrived training observations are learned to modify the model of the SLFNs. As soon as the learning procedure for the arrived observations is completed, the data is discarded. Moreover, it has no prior knowledge about the amount of the observations which will be presented. Therefore, OS-ELM is an elegant sequential learning algorithm which can handle both the RBF and additive nodes in the same framework and can be used to both the classification and function regression problems. OS-ELM proves to be a very fast and accurate online sequential learning algorithm[9-11], which can provide better generalization performance in faster speed compared with other sequential learning algorithms such as GAP-RBF, GGAP-RBF, SGBP, RAN, RANEKF and MRAN etc.

However, various data sets require more robust learning algorithms. Due to the random generation of the parameters for the hidden nodes, the robustness and stability of OS-ELM sometimes cannot be guaranteed, similar to ELM. Some ensemble based methods and pruning based methods have been applied to ELM to improve its robustness[12-15]. Ensemble learning is a learning scheme where a collection of a finite number of learners is trained for the same task[16-17]. It has been demonstrated that the generalization ability of a learner can be significantly improved by ensembling a set of learners. In [18] a simple ensemble OS-ELM, i.e., EOS-ELM, has been investigated. However, Zhou et al.[19] proved that selective ensemble is better a choice. We apply this idea to OS-ELM. At first, a novel selective ensemble algorithm--PSOSEN, is proposed. PSOSEN adopts particle swarm optimization[20] to select the individual OS-ELMs to form the ensemble. It should be noted that PSOSEN is a general selective ensemble algorithm suitable for any learning algorithms.

Different from batch learning, online learning algorithms need to perform learning continually. Therefore the complexity of the learning algorithm should be taken into account. Obviously, performing selective ensemble learning each step is not a good choice for sequential learning. Thus we designed an adaptive selective ensemble framework for OS-ELM. A set of OS-ELMs are trained online, and the root mean square error (RMSE) will always be calculated. The error will be compared with a pre-set threshold $\lambda$. If RMSE is bigger than the threshold, it means the model is not accurate. Then PSOSEN will be performed and a selective ensemble $M$ is obtained. Otherwise, it means the model is relatively accurate and the ensemble will not be selected. Then the output of the system is calculated as the average output of the individuals in the ensemble set. And each individual OS-ELM will be updated recursively.

UCI data sets[21], which contain both regression and classification data, are used to verify the feasibility of the proposed ROS-ELM algorithm. Comparisons of three aspects including RMSE, standard deviation and running time between OS-ELM, EOS-ELM and ROS-ELM are presented. The results convincingly show that ROS-ELM significantly improves the robustness and stability compared with OS-ELM and EOS-ELM.

The rest of the paper is organized as follows: In section 2, previous work including ELM and OS-ELM are reviewed. A novel selective ensemble based on particle swarm optimization is presented in section 3. An adaptive selective ensemble framework is designed for OS-ELM referred to as ROS-ELM, is proposed in section 4. Experiments are carried out in section 4 and the comparison results are also presented. In section 5, we draw the conclusion of the paper.

## 2  Review of related works

In this section, both the basic ELM algorithm and the online version OS-ELM are reviewed in brief as the background knowledge for our work.

### 2.1  Extreme Learning Machine (ELM)

ELM algorithm is derived from single hidden layer feedforward neural networks (SLFNs). Unlike traditional SLFNs, ELM assigns the parameters of the hidden nodes randomly without any iterative tuning. Besides, all the parameters of the hidden nodes in ELM are independent with each other. Hence ELM can be seen as generalized SLFNs. The only problem for ELM is to calculate the output weights.

Given $N$ training samples $(x_i, t_i) \in R^n \times R^m$, where $x_i$ is an input vector of $n$ dimensions and $t_i$ is a target vector of $m$ dimensions. Then SLFNs with $\tilde{N}$ hidden nodes each with output function $G(a_i, b_i, x)$ are mathematically modeled as

$$f_{\tilde{N}}(x_j) = \sum_{i=1}^{\tilde{N}} \beta_i G(a_i, b_i, x_j) = t_j, j = 1, \cdots, N. \qquad (1)$$

Where $(a_i, b_i)$ are parameters of hidden nodes, and $\beta_i$ is the weight vector connecting the $i$ th hidden node and the output node. To simplify, equation (1) can be written equivalently as:

$$H\beta = T \qquad (2)$$

where

$$H(a_1,\cdots,a_N,b_1,\cdots,b_{\tilde{N}},x_1,\cdots,x_N) = \begin{bmatrix} G(a_1,b_1,x_1) & \cdots & G(a_{\tilde{N}},b_{\tilde{N}},x_1) \\ \vdots & \ddots & \vdots \\ G(a_1,b_1,x_N) & \cdots & G(a_{\tilde{N}},b_{\tilde{N}},x_N) \end{bmatrix}_{N \times \tilde{N}}$$

(3)

$$\beta = \begin{bmatrix} \beta_1^T \\ \vdots \\ \beta_{\tilde{N}}^T \end{bmatrix}_{\tilde{N} \times m} \qquad T = \begin{bmatrix} t_1^T \\ \vdots \\ t_N^T \end{bmatrix}_{N \times m}$$

(4)

$H$ is called the hidden layer output matrix of the neural network, and the $i$ th column of $H$ is the output of the $i$ th hidden node with respect to inputs $x_1, x_2, \cdots, x_N$.

In ELM, $H$ can be easily obtained as long as the training set is available and the parameters $(a_i, b_i)$ are randomly assigned. Then ELM evolves into a linear system and the output weights $\beta$ are calculated as:

$$\hat{\beta} = H^{\dagger}T$$

(5)

where $H^{\dagger}$ is the Moore-Penrose generalized inverse of matrix $H$.

The ELM algorithm can be summarized in three steps as shown in Algorithm 1:

---

Algorithm 1

Input:

A training set $\aleph = \{(x_i, t_i) | x_i \in R^n, t_i \in R^m, i = 1, \cdots, N\}$, hidden node output function $G(a_i, b_i, x)$, and the number of hidden nodes $\tilde{N}$

Steps:

1. Assign parameters of hidden nodes $(a_i, b_i)$ randomly, $i = 1, \cdots, \tilde{N}$.
2. Calculate the hidden layer output matrix $H$.
3. Calculate the output weight $\beta$: $\hat{\beta} = H^{\dagger}T$, where $H^{\dagger}$ is the Moore-Penrose generalized inverse of hidden layer output matrix $H$.

---

### 2.2 OS-ELM

In many industrial applications, it is impossible to have all the training data available before the learning process. It is common that the training observations are sequentially inputted to the learning algorithm, i.e., the observations arrive one-by-one or chunk-by-chunk. In this case, the batch ELM algorithm is no longer applicable. Hence, a fast and accurate online sequential extreme learning machine was proposed to deal with online learning.

The output weight $\beta$ obtained from equation (5) is actually a least-squares solution of equation (2). Given $rank(H) = \tilde{N}$, the number of hidden nodes, $H^{\dagger}$ can be presented as:

$$H^{\dagger} = \left(H^T H\right)^{-1} H^T \qquad (6)$$

This can also be called the left pseudoinverse of $H$ for it satisfies the equation $H^{\dagger}H = I_{\tilde{N}}$. If $H^T H$ tends to be singular, smaller network size $\tilde{N}$ and larger data number $N_0$ should be chosen in the initialization step of OS-ELM. Substituting equation (6) to equation (5), we can get

$$\hat{\beta} = \left(H^T H\right)^{-1} H^T T \tag{7}$$

which is the least-squares solution to equation (2). Then the OS-ELM algorithm can be deduced by recursive implementation of the least-squares solution of (7).

There are two main steps in OS-ELM, initialization step and update step. In the initialization step, the number of training data $N_0$ needed in this step should be equal to or larger than network size $\tilde{N}$. In the update step, the learning model is updated with the method of recursive least square (RLS). And only the newly arrived single or chunk training observations are learned, which will be discarded as soon as the learning step is completed.

The two steps for OS-ELM algorithm in general:

a. Initialization step: batch ELM is used to initialize the learning system with a small chunk of initial training data $\aleph_0 = \{(x_i, t_i)\}_{i=1}^{N_0}$ from given training set $\aleph = \{(x_i, t_i), x_i \in R^n, t_i \in R^m, i = 1, \cdots\}$ $N_0 \geq \tilde{N}$.

1. Assign random input weights $a_i$ and bias $b_i$ (for additive hidden nodes) or center $a_i$ and impact factor $b_i$ (for RBF hidden nodes), $i = 1, \cdots, \tilde{N}$.

2. Calculate the initial hidden layer output matrix:

$$H_0 = \begin{bmatrix} G(a_1, b_1, x_1) & \cdots & G(a_{\tilde{N}}, b_{\tilde{N}}, x_1) \\ \vdots & \ddots & \vdots \\ G(a_1, b_1, x_{N_0}) & \cdots & G(a_{\tilde{N}}, b_{\tilde{N}}, x_{N_0}) \end{bmatrix}_{N_0 \times \tilde{N}} \tag{8}$$

3. Calculate the initial output weight $\beta^{(0)} = P_0 H_0^T T_0$, where $P_0 = \left(H_0^T H_0\right)^{-1}$ and $T_0 = [t_1, \cdots, t_{N0}]^T$.

4. Set k=0. Initialization is finished.

b. Sequential learning step:

The $(k+1)$ th chunk of new observations can be expressed as:

$$\aleph_{k+1} = \{(x_i, t_i)\}_{i=\left(\sum_{j=0}^{k} N_j\right)+1}^{\sum_{j=0}^{k+1} N_j} \tag{9}$$

where $\aleph_{k+1}$ represents the number of observations in the (k+1)th chunk newly arrived.

1. Compute the partial hidden layer output matrix $H_{k+1}$ for the $(k+1)$ th chunk.

$$H_{k+1} = \begin{bmatrix} G\left(a_1, b_1, x_{\left(\sum_{j=0}^{k} N_j\right)+1}\right) & \cdots & G\left(a_{\tilde{N}}, b_{\tilde{N}}, x_{\left(\sum_{j=0}^{k} N_j\right)+1}\right) \\ \vdots & \ddots & \vdots \\ G\left(a_1, b_1, x_{\sum_{j=0}^{k+1} N_j}\right) & \cdots & G\left(a_{\tilde{N}}, b_{\tilde{N}}, x_{\sum_{j=0}^{k+1} N_j}\right) \end{bmatrix}_{N_{k+1} \times \tilde{N}} \tag{10}$$

2. Set $T_{k+1} = \left[t_{\left(\sum_{j=0}^{k} N_j\right)+1}, \cdots, t_{\sum_{j=0}^{k+1} N_j}\right]^T$. And we have

$$K_{k+1} = K_k + H_{k+1}^T H_{k+1} \tag{11}$$

$$\beta^{(k+1)} = \beta^{(k)} + K_{k+1}^{-1} H_{k+1}^T \left(T_{k+1} - H_{k+1}\beta^{(k)}\right) \tag{12}$$

To avoid calculating inverse in the iterative procedure, $K_{k+1}^{-1}$ is factored as the following according to Woodbury formula:

$$K_{k+1}^{-1} = \left(K_k + H_{k+1}^T H_{k+1}\right)^{-1} \\ = K_k^{-1} - K_k^{-1} H_{k+1}^T \left(I + H_{k+1} K_k^{-1} H_{k+1}^T\right)^{-1} H_{k+1} K_k^{-1} \quad (13)$$

Let $P_{k+1} = K_{k+1}^{-1}$.

3. Calculate the output weight $\beta^{(k+1)}$, according to the updating equations:

$$P_{k+1} = P_k - P_k H_{k+1}^T \left(I + H_{k+1} P_k H_{k+1}^T\right)^{-1} H_{k+1} P_k \quad (14)$$

$$\beta^{(k+1)} = \beta^{(k)} + P_{k+1} H_{k+1}^T \left(T_{k+1} - H_{k+1} \beta^{(k)}\right) \quad (15)$$

4. Set $k = k+1$. Go to step b.

## 3　Particle Swarm Optimization Selective Ensemble

In this section, a novel selective ensemble method referred to as particle swarm optimization selective ensemble (PSOSEN) is proposed. PSOSEN adopts particle swarm optimization to select the good learners and combine their predictions. Detailed procedures of the PSOSEN algorithm will be introduced in this section.

Zhou et al.[19] have demonstrated that ensembling many of the available learners may be better than ensembling all of those learners in both regression and classification. The detailed proof of this conclusion will not be presented in this paper. However, one important problem for selective ensemble is how to select the good learners in a set of available learners.

The novel approach--PSOSEN, is proposed to select good learners in the ensemble. PSOSEN is based on the idea of heuristics. It assumes each learner can be assigned a weight, which could characterize the fitness of including this learner in the ensemble.

Then the learner with the weight bigger than a pre-set threshold $\lambda$ could be selected to join the ensemble.

We will explain the principle of PSOSEN from the context of regression. We use $\omega_i$ to denote the weight of the $i$ th component learner. The weight should satisfy the following equations:

$$0 \leq \omega_i \leq 1 \tag{16}$$

$$\sum_{i=1}^{N} \omega_i = 1 \tag{17}$$

Then the weight vector is:

$$\omega = (\omega_1, \omega_2, ..., \omega_N) \tag{18}$$

Suppose input variables $x \in R^m$ according to the distribution $p(x)$, the true output of $x$ is $d(x)$, and the actual output of the $i$ th learner is $f_i(x)$. Then the output of the simple weighted ensemble on $x$ is:

$$\hat{f}(x) = \sum_{i=1}^{N} \omega_i f_i(x) \tag{19}$$

Then the generalization error $E_i(x)$ of the $i$ th learner and the generalization error $\hat{E}(x)$ of the ensemble are calculated on $x$ respectively:

$$E_i(x) = (f_i(x) - d(x))^2 \tag{20}$$

$$\hat{E}(x) = (\hat{f}(x) - d(x))^2 \tag{21}$$

The generalization error $E_i$ of the $i$ th learner and that of the ensemble $\hat{E}$ are calculated on $p(x)$ respectively:

$$E_i = \int dx p(x) E_i(x) \tag{22}$$

$$\hat{E} = \int dx p(x) \hat{E}(x) \tag{23}$$

We then define the correlation between the $i$ th and the $j$ th component learner as following:

$$C_{ij} = \int dx p(x)(f_i(x) - d(x))(f_j(x) - d(x)) \tag{24}$$

Obviously $C_{ij}$ satisfies the following equations:

$$C_{ii} = E_i \tag{25}$$

$$C_{ij} = C_{ji} \tag{26}$$

Considering the equations defined above, we can get:

$$\hat{E}(x) = (\sum_{i=1}^{N} \omega_i f_i(x) - d(x))(\sum_{j=1}^{N} \omega_j f_j(x) - d(x)) \tag{27}$$

$$\hat{E} = \sum_{i=1}^{N} \sum_{j=1}^{N} \omega_i \omega_j C_{ij} \tag{28}$$

To minimize the generalization error of the ensemble, according to equation (28), the optimum weight vector can be obtained as:

$$\omega_{opt} = \arg\min_{\omega}(\sum_{i=1}^{N}\sum_{j=1}^{N}\omega_i\omega_j C_{ij}) \qquad (29)$$

The $k$ th variable of $\omega_{opt}$, i.e., $\omega_{opt.k}$, can be solved by Lagrange multiplier:

$$\frac{\partial\ (\sum_{i=1}^{N}\sum_{j=1}^{N}\omega_i\omega_j C_{ij} - 2*\lambda(\sum_{i=1}^{N}\omega_i - 1)\ )}{\partial \omega_{opt.k}} = 0 \qquad (30)$$

The equation can be simplified to:

$$\sum_{\substack{j=1\\j\neq k}}^{N}\omega_{opt.k}C_{kj} = \lambda \qquad (31)$$

Taking equation (2) into account, we can get:

$$\omega_{opt.k} = \frac{\sum_{j=1}^{N}C_{kj}^{-1}}{\sum_{i=1}^{N}\sum_{j=1}^{N}\omega_i\omega_j C_{ij}^{-1}} \qquad (32)$$

Equation (32) gives the direct solution for $\omega_{opt}$. But the solution seldom work well in real word applications. Due to the fact that some learners are quite similar in performance, when a number of learners are available, the correlation matrix $C_{ij}$ may be irreversible or ill-conditioned.

Although we cannot obtain the optimum weights of the learner directly, we can approximate them in some way. Equation (29) can be viewed as an optimization problem. As particle swarm optimization has been proved to be a powerful optimization tool, PSOSEN is then proposed. The basic PSO algorithm is showed in Figure 1.

PSOSEN randomly assigns a weight to each of the available learners at first. Then it employs particle swarm optimization algorithm to evolve those weights so that the weights can characterize the fitness of the learners in joining the ensemble. Finally, learners whose weight is bigger than a pre-set threshold $\lambda$ are selected to form the ensemble. Note that if all the evolved weights are bigger than the threshold $\lambda$, then all the learners will be selected to join the ensemble.

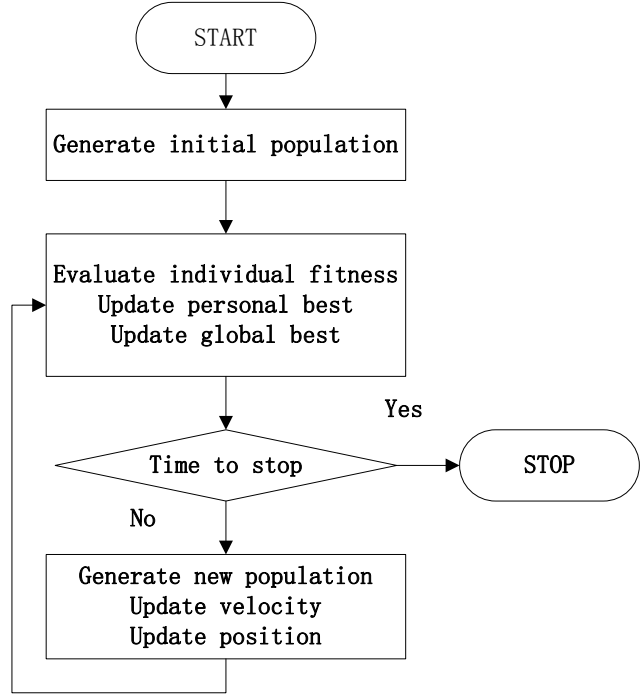

**Fig. 1.** Flowchart for particle swarm optimization algorithm.

PSOSEN can be applied to both regression and classification problems for the purpose of the weights evolving process is only to select the component learners. In particular, the output of the ensemble for regression are combined via simple averaging instead of weighted averaging. The reason is that previous work [19] showed that using the weights both in selection of the component learners and combination of the outputs tends to suffer the overfitting problem.

In the process of generating population, the goodness of the individuals are eva-luated via validation data bootstrap sampled from the training data set. We use $\hat{E}_\omega^V$ to denote the generalization error of the ensemble, which corresponds to individual $\omega$ on the validation data $V$. Obviously $\hat{E}_\omega^V$ can describe the goodness of $\omega$. The smaller $\hat{E}_\omega^V$ is, the better $\omega$ is. So, PSOSEN adopts $f(\omega) = 1/\hat{E}_\omega^V$ as the fitness function.

The PSOSEN algorithm is summarized as follows. $S_1, S_2, ..., S_T$ are bootstrap samples generated from original training data set. A component learner $N_t$ is trained from each $S_T$. And an selective ensemble $N^*$ is built from $N_1, N_2, ..., N_T$. The output is the average output of the ensemble for regression, or the class label who receives the most number in voting process for classification.

**PSOSEN**

**Input: training set S, learner L, trial T, threshold $\lambda$**
**Steps:**
1. for t = 1 to T{
   $S_T$ = bootstrap sample from S
   $N_T = L(S_T)$
}
2. generate a population of weight vectors
3. evolve the population by PSO, where the fitness of the weight vector $\omega$ is defined as $f(\omega) = 1/\hat{E}_\omega^V$.
4. $\omega^*$ = the evolved best weight vector

Output: ensemble $N^*$:
$$N^*(x) = Ave \sum_{\omega_t^* > \lambda} N_t(x) \quad \text{for regression}$$
$$N^*(x) = \arg\max_{y \in Y} \sum_{\omega_t^* > \lambda, N_t(x) = y} 1 \quad \text{for classification}$$

## 4  Robust online sequential extreme learning machine

In this section, the detailed procedure of the proposed robust online sequential learning algorithm is introduced. The novel selective ensemble algorithm--PSOSEN is applied to the original OS-ELM to improve the robustness. In order to reduce the complexity and employ PSOSEN flexibly, an adaptive framework is then designed. The new algorithm, which is based on OS-ELM and adaptive ensemble, is termed as robust online sequential extreme learning machine (ROS-ELM).

The flowchart of ROS-ELM is showed as follows:

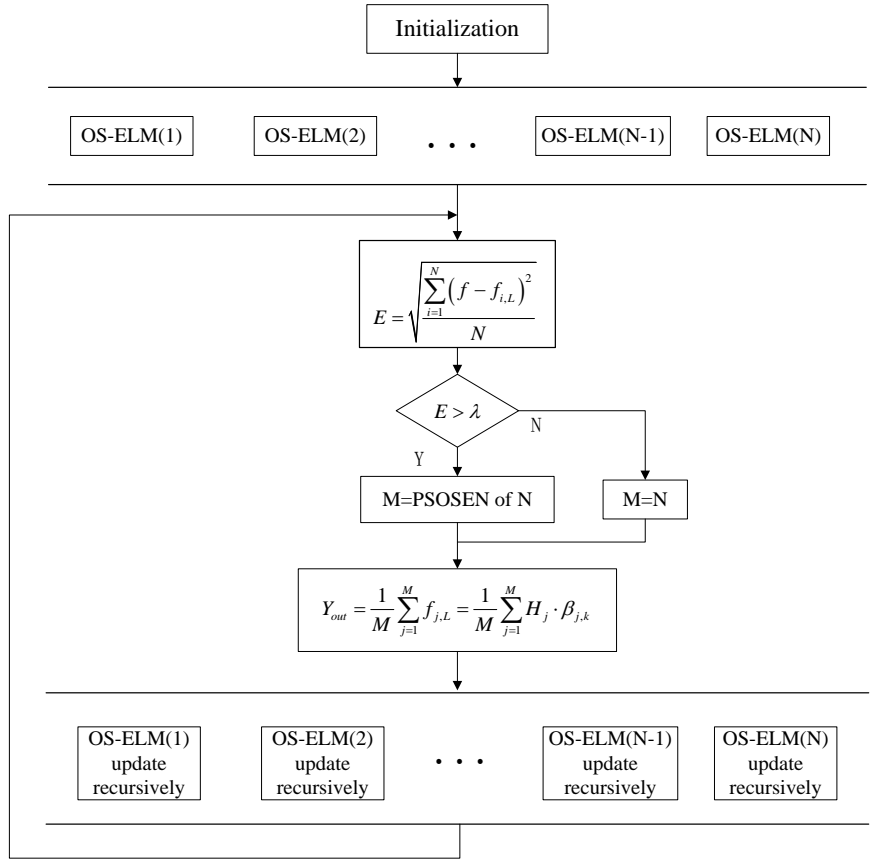

**Fig. 2.** Flowchart for the ROS-ELM algorithm.

Online sequential learning is necessary in many industrial applications. In this situations, training data can only be obtained sequentially. Although OS-ELM is proposed as a fast and accurate online learning algorithm, it still suffers from the robustness problem, which results from the random generation of the input weights and biases, similar to ELM. Ensemble methods has been investigated in OS-ELM, i.e., the EOS-ELM algorithm[18]. However, it is only very simple ensemble method, which just calculates the average of all the N individual OS-ELMs. In this section, selective ensemble, which is superior to simple ensemble, is adopted to OS-ELM. The novel selective ensemble method--PSOSEN, proposed in section 3, is chose as the algorithm. Apparently, performing PSOSEN each step is a time consuming process. We design an adaptive framework to determine whether to perform PSOSEN or simple ensemble. Thus the robustness and the complexity can be balanced well. The ROS-ELM algorithm can be explained as follows:

First, N individual OS-ELMs are initialized. The number of nodes is same for each OS-ELM. While the input weights and biases for each OS-ELM are randomly generated.

Second, the RMSE error is calculated:

$$E = \sqrt{\frac{\sum_{i=1}^{N}(f - f_{i,L})^2}{N}} \qquad (33)$$

where $f$ is the expected output, while $f_{i,L}$ is the actual output of the $i$ th individual OS-ELM.

The RMSE will be compared with a pre-set threshold $\lambda$. If $E$ is bigger than $\lambda$, which means simple ensemble is not accurate, PSOSEN is performed and a selective ensemble $M$ is obtained. And if $E$ is smaller than $\lambda$, which indicates that simple ensemble is relatively accurate, the ensemble will not be selected.

Third, the output of the system is calculated as the average output of the individual in the ensemble set:

$$Y_{out} = \frac{1}{M}\sum_{j=1}^{M} f_{j,L} = \frac{1}{M}\sum_{j=1}^{M} H_j \cdot \beta_{j,k} \qquad (34)$$

where $H_j$ is the output matrix of the $j$ th OS-ELM, and $\beta_{j,k}$ is the output weight calculated by the $j$ th OS-ELM at step $k$.

At last, each OS-ELM will update recursively according to the update equations presented in section 2.

## 5 Performance evaluation of ROS-ELM

In this section, a series of experiments were conducted to evaluate the performance of the proposed ROS-ELM algorithm. OS-ELM and EOS-ELM are also compared with ROS-ELM in this section. All the experiments were carried out in the MatlabR2012b environment on a desktop of CPU 3.40GHz and 8GB RAM.

### 5.1 Model selection

For OS-ELM, the number of hidden nodes is the only parameter needs to be determined. Cross-validation method are usually used to choose this parameter. Fifty trials of simulations are performed respectively for classification, regression and time-series problems. The number of hidden nodes is then determined by the validation error.

For EOS-ELM and ROS-ELM, there is another parameter that needs to be determined, i.e., the number of networks in the ensemble. The parameter is set from 5 to 30 with the interval 5. Finally, the optimal parameter is selected according to the RMSE for regression, testing accuracy for classification and standard deviation value. Under the same problem, the number of OS-ELMs is selected based on the lowest standard deviation and the comparable RMSE or accuracy compared with OS-ELM. Table 1 is an example of selecting the optimal number of networks for ROS-ELM with RBF hidden nodes on New-thyroid dataset. As illustrated by Table 1, the lowest standard deviation occurs when the number of OS-ELMs is 20. Meanwhile, the prediction accuracy of ROS-ELM is better than OS-ELM. Hence we set the number of networks to be 20 for the New-thyroid dataset. The numbers of OS-ELMs for other datasets are determined in the same way.

Both the Gaussian radial basis function (RBF) $G(a,b,x) = \exp(-\|x-a\|^2 / b)$ and the sigmoid additive $G(a,b,x) = 1/(1+\exp(-(a \cdot x + b)))$ are adopted as activation function in OS-ELM, EOS-ELM and ROS-ELM.

**Table 1.** Network selection for New-thyroid dataset.

| Num of networks | 1 | 5 | 10 | 15 | 20 | 25 | 30 |
|---|---|---|---|---|---|---|---|
| Testing accuracy | 90.73 | 91.25 | 90.65 | 90.18 | 92.24 | 91.79 | 91.8 |
| Testing Dev | 0.0745 | 0.0254 | 0.0316 | 0.0276 | 0.0138 | 0.024 | 0.0156 |

In the experiments, OS-ELM and EOS-ELM were compared with ROS-ELM. Some general information of the benchmark datasets used in our evaluations is listed in Table 2. Both regression and classification problems are included.

**Table 2.** Specification of benchmark datasets.

|  | Dataset | Classes | Training data | Testing data | Attributes |
|---|---|---|---|---|---|
| Regression problems | Auto-MPG | - | 320 | 72 | 7 |
|  | Abalone | - | 3000 | 1177 | 8 |
|  | California housing | - | 8000 | 12640 | 8 |
|  | Mackey-Glass | - | 4000 | 500 | 4 |
| Classification problems | Zoo | 7 | 71 | 30 | 17 |
|  | Wine | 3 | 120 | 58 | 13 |
|  | New-thyroid | 3 | 140 | 75 | 5 |
|  | Monks-1 | 2 | 300 | 132 | 6 |
|  | Image segmentation | 7 | 1500 | 810 | 19 |
|  | Satellite image | 6 | 4435 | 2000 | 36 |

For OS-ELM, the input weights and biases with additive activation function or the centers with RBF activation function were all generated from the range [-1, 1]. For regression problems, all the inputs and outputs were normalized into the range [0, 1], while the inputs and outputs were normalized into the range [-1, 1] for classification problems.

The benchmark datasets studied in the experiments are from UCI Machine Learning Repository except California Housing dataset from the StatLib Repository. Besides, a time-series problem, Mackey-Glass, from UCI was also adopted to test our algorithms.

### 5.2 Algorithm evaluation

To verify the superiority of the ROS-ELM, RMSE for regression problems and testing accuracy for classification problems are respectively computed. The evaluation results are presented in Table 3 and Table 4, which are respectively corresponding to the models with sigmoid hidden nodes and RBF hidden nodes. Each result is an average of 50 trials. And in every trial of one problem, the training and testing samples were randomly adopted from the dataset that was addressed currently.

From the comparison results of Table 2 and Table 3, we can easily find that ROS-ELM and EOS-ELM are more time consuming than OS-ELM, but they still keep relatively fast speed at most of the time. What's important, ROS-ELM and EOS-ELM attain lower testing deviation and more accurate regression or classification results than OS-ELM. In terms of the comparison between ROS-ELM and EOS-ELM, it can be observed that ROS-ELM takes a little more time than EOS-ELM, which results from the selective ensemble by PSOSEN in ROS-ELM instead of simply averaging the networks in EOS-ELM. It should be noted that the complexity of ROS-ELM is adjustable, which depends on the threshold $\lambda$. Nevertheless, ROS-ELM always outperforms EOS-ELM in terms of accuracy and testing deviation. Hence with the adaptive ensemble framework, ROS-ELM tends to generate more accurate and robust results.

To verify the reliability of the proposed ROS-ELM more convincingly, an artificial dataset is dissected for instance. The dataset was generated from the function $y = x^2 + 3x + 2$, comprising 4500 training data and 1000 testing data. Figure 3 and 4 explicitly depict the variability of training accuracy of ROS-ELM, EOS-ELM and OS-ELM with respect to the number of training data in the process of learning. It can be observed that with the increasing number of training samples, RMSE values of the three methods significantly decline. As the online learning progressed, the training models are continuously updated and corrected. We can then conclude that the more training data the system learns, the more precise the model is. Whether sigmoid or RBF the hidden nodes is, ROS-ELM always obtains smaller RMSE than EOS-ELM and OS-ELM, which indicates that the performance of ROS-ELM is considerably accurate and robust compared with the other methods. Moreover, the smaller testing dev of ROS-ELM in Table 3 and 4 also confirms the robust performance of ROS-ELM.

**Table 3.** Comparison of OS-ELM, EOS-ELM and ROS-ELM for sigmoid hidden nodes.

| Datasets | Algorithm | #Nodes | #Network | Training time (s) | RMSE or Accuracy | | Testing Dev |
|---|---|---|---|---|---|---|---|
| | | | | | Training RMSE | Testing RMSE | |
| Au-to-MPG | OS-ELM | 25 | | 0.0121 | 0.0695 | 0.0745 | 0.0087 |
| | EOS-ELM | 25 | 20 | 0.2385 | 0.0691 | 0.0751 | 0.0065 |
| | ROS-ELM | 25 | 20 | 1.9083 | 0.0683 | 0.0741 | 0.0053 |
| Abalone | OS-ELM | 25 | | 0.1191 | 0.0758 | 0.0782 | 0.0049 |
| | EOS-ELM | 25 | 5 | 0.5942 | 0.0754 | 0.0775 | 0.0023 |
| | ROS-ELM | 25 | 5 | 4.1528 | 0.0742 | 0.0758 | 0.0015 |
| Mack-ey-Glass | OS-ELM | 120 | | 0.9827 | 0.0177 | 0.0185 | 0.0018 |
| | EOS-ELM | 120 | 5 | 4.8062 | 0.0176 | 0.0183 | 0.0007 |
| | ROS-ELM | 120 | 5 | 25.1608 | 0.0173 | 0.0179 | 0.0006 |
| California Housing | OS-ELM | 50 | | 0.6871 | 0.1276 | 0.1335 | 0.0035 |
| | EOS-ELM | 50 | 5 | 3.2356 | 0.1280 | 0.1337 | 0.0019 |
| | ROS-ELM | 50 | 5 | 15.6326 | 0.1238 | 0.1323 | 0.0014 |
| Zoo | OS-ELM | 35 | | 0.0042 | 100% | 93.09% | 0.0498 |
| | EOS-ELM | 35 | 25 | 0.0986 | 100% | 93.68% | 0.0375 |
| | ROS-ELM | 35 | 25 | 0.8768 | 100% | 94.51% | 0.0315 |
| Wine | OS-ELM | 30 | | 0.0053 | 99.83% | 97.24% | 0.0251 |
| | EOS-ELM | 30 | 5 | 0.0247 | 99.79% | 97.49% | 0.0117 |
| | ROS-ELM | 30 | 5 | 0.1628 | 99.88% | 98.01% | 0.0094 |
| New-thyroid | OS-ELM | 20 | | 0.0043 | 93.18% | 89.66% | 0.1138 |
| | EOS-ELM | 20 | 15 | 0.0627 | 94.32% | 90.92% | 0.02765 |
| | ROS-ELM | 20 | 15 | 0.5012 | 95.23% | 91.78% | 0.01986 |
| Monks-1 | OS-ELM | 80 | | 0.0378 | 89.34% | 78.77% | 0.0325 |
| | EOS-ELM | 80 | 15 | 0.5432 | 89.18% | 78.79% | 0.0187 |
| | ROS-ELM | 80 | 15 | 4.2804 | 90.24% | 79.85% | 0.0138 |
| Image segmen-tation | OS-ELM | 180 | | 1.8432 | 97.07% | 94.83% | 0.0078 |
| | EOS-ELM | 180 | 20 | 36.2458 | 97.08% | 94.79% | 0.0055 |
| | ROS-ELM | 180 | 20 | 254.0721 | 97.56% | 95.21% | 0.0043 |
| Satellite image | OS-ELM | 400 | | 42.2503 | 92.82% | 88.92% | 0.0058 |
| | EOS-ELM | 400 | 20 | 853.2675 | 92.80% | 89.05% | 0.0026 |
| | ROS-ELM | 400 | 20 | 6928.0968 | 93.96% | 90.16% | 0.0018 |

**Table 4.** Comparison of OS-ELM, EOS-ELM and ROS-ELM for RBF hidden nodes.

| Datasets | Algorithm | #Nodes | #Network | Training time (s) | RMSE or Accuracy | | Testing Dev |
|---|---|---|---|---|---|---|---|
| | | | | | Training RMSE | Testing RMSE | |
| Au-to-MPG | OS-ELM | 25 | | 0.0302 | 0.0685 | 0.0763 | 0.0081 |
| | EOS-ELM | 25 | 20 | 0.5986 | 0.0681 | 0.0754 | 0.0072 |
| | ROS-ELM | 25 | 20 | 4.1862 | 0.0672 | 0.0741 | 0.0063 |
| Abalone | OS-ELM | 25 | | 0.3445 | 0.0753 | 0.0775 | 0.0027 |
| | EOS-ELM | 25 | 25 | 8.5762 | 0.0752 | 0.0773 | 0.0023 |
| | ROS-ELM | 25 | 25 | 49.3562 | 0.0741 | 0.0761 | 0.0017 |
| Mack-ey-Glass | OS-ELM | 120 | | 1.6854 | 0.0181 | 0.0185 | 0.0092 |
| | EOS-ELM | 120 | 5 | 8.4304 | 0.0171 | 0,0171 | 0.0028 |
| | ROS-ELM | 120 | 5 | 55.1469 | 0.0159 | 0.0156 | 0.0016 |
| California Housing | OS-ELM | 50 | | 1.8329 | 0.1298 | 0.1317 | 0.0017 |
| | EOS-ELM | 50 | 5 | 9.0726 | 0.1296 | 0.1316 | 0.0011 |
| | ROS-ELM | 50 | 5 | 64.9625 | 0.1202 | 0.1243 | 0.0009 |
| Zoo | OS-ELM | 35 | | 0.0074 | 99.91% | 91.15% | 0.0508 |
| | EOS-ELM | 35 | 15 | 0.1028 | 99.87% | 90.47% | 0.0429 |
| | ROS-ELM | 35 | 15 | 0.8543 | 99.93% | 91.26% | 0.0315 |
| Wine | OS-ELM | 30 | | 0.0132 | 99.73% | 97.09% | 0.0225 |
| | EOS-ELM | 30 | 5 | 0.6015 | 99.76% | 97.18% | 0.0138 |
| | ROS-ELM | 30 | 5 | 4.9028 | 99.84% | 98.14% | 0.0087 |
| New-thyroid | OS-ELM | 20 | | 0.0118 | 93.45% | 89.92% | 0.0702 |
| | EOS-ELM | 20 | 15 | 0.1682 | 93.87% | 89.86% | 0.0428 |
| | ROS-ELM | 20 | 15 | 1.2315 | 94.68% | 91.02% | 0.0315 |
| Monks-1 | OS-ELM | 80 | | 0.1024 | 94,58% | 87.28% | 0.0882 |
| | EOS-ELM | 80 | 20 | 2.1567 | 93.69% | 86.34% | 0.0324 |
| | ROS-ELM | 80 | 20 | 15.2896 | 95.71% | 88.47% | 0.0195 |
| Image segmen-tation | OS-ELM | 180 | | 2.6702 | 94.98% | 91.92% | 0.0324 |
| | EOS-ELM | 180 | 5 | 13.2174 | 94.39% | 91.35% | 0.0148 |
| | ROS-ELM | 180 | 5 | 90.2856 | 96.02% | 953.24% | 0.0079 |
| Satellite image | OS-ELM | 400 | | 45.2702 | 93.62% | 89.54% | 0.0056 |
| | EOS-ELM | 400 | 10 | 448.1347 | 93.86% | 89.37% | 0.0034 |
| | ROS-ELM | 400 | 10 | 3145.8528 | 94.75% | 90.48% | 0.0019 |

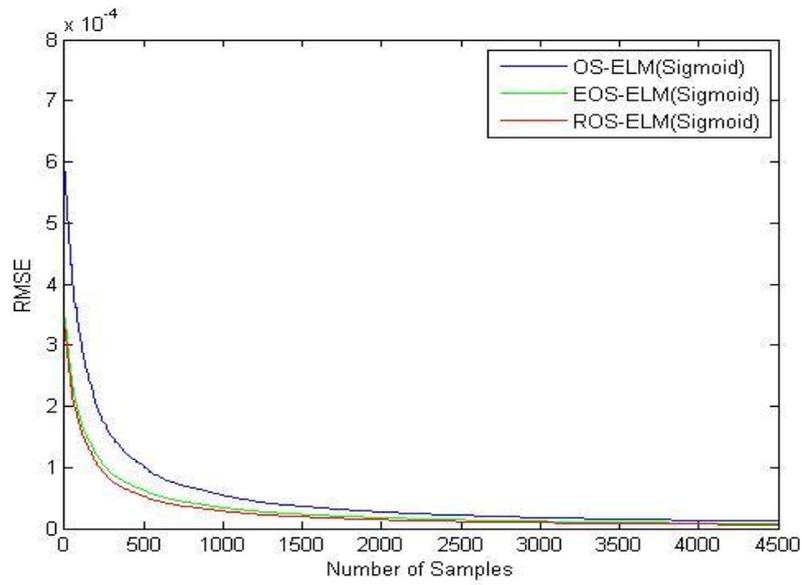

**Fig. 3.** RMSE with respect to the number of training samples for sigmoid hidden nodes.

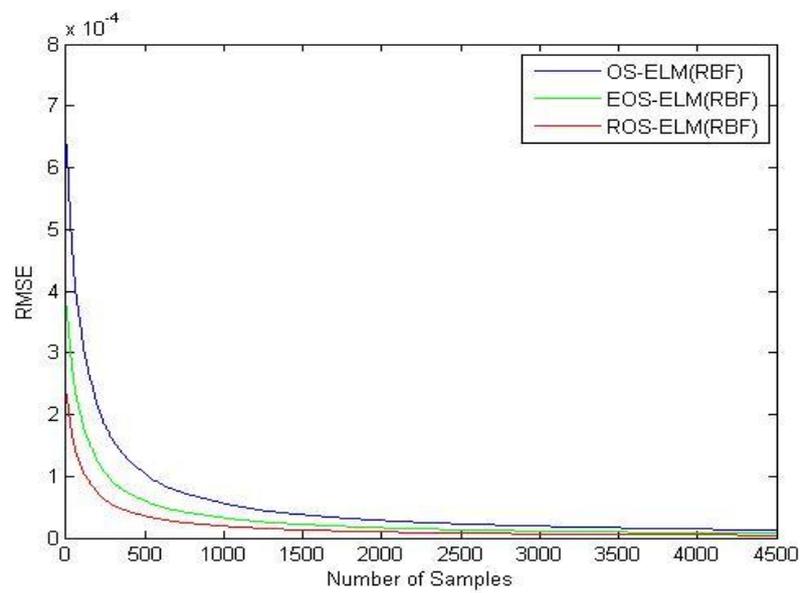

**Fig. 4.** RMSE with respect to the number of training samples for RBF hidden nodes.

Hence, by analyzing the results in Figure 3, Figure 4, Table 3 and Table 4 comprehensively, we can draw the conclusion that ROS-ELM improves the accuracy and robustness of the online sequential learning algorithm significantly for both regression and classification applications, with a still relative fast speed.

## 6  Conclusion

In this paper, a robust online sequential extreme learning machine algorithm is proposed. To improve the robustness and stability of the sequential learning algorithm, we apply the selective ensemble method to OS-ELM. And in purpose of balancing the complexity and accuracy, an adaptive selective ensemble framework for OS-ELM is designed, which is referred to as ROS-ELM. In addition, before building the ROS-ELM system, a novel selective ensemble algorithm is proposed which is suitable for any learning methods, both batch learning and sequential learning. The proposed selective ensemble algorithm-PSOSEN, adopts particle swarm optimization method to select individual learner to form the ensemble. Experiments were carried out on UCI data set. The results convincingly show that ROS-ELM improves the robustness and stability of OS-ELM, while also keeps balance on complexity.